\documentclass{article}
\usepackage[utf8]{inputenc}
\usepackage{multicol} 
\usepackage[left=3cm, right=2.5cm, top=1.5cm, bottom=2.5cm]{geometry}
\usepackage{wrapfig} 
\usepackage{graphicx}
\usepackage{subfigure} 
\usepackage{multirow}
\usepackage{amsmath}
\usepackage{booktabs}
\usepackage{amssymb}
\usepackage{hyperref}
\usepackage{etoolbox} 
\usepackage{yax}      
\usepackage{float}
\usepackage[para,online,flushleft]{threeparttable}
\usepackage{authblk} 
\usepackage[numbers]{natbib} 

\hypersetup{
    colorlinks=true,
    linkcolor=blue,
    filecolor=blue,
    urlcolor=blue,
    citecolor=cyan,
}

\providecommand{\keywords}[1]
{
  \small	
  \textbf{\textit{Keywords:}} #1
}

\title{Ensemble architecture in polyp segmentation}
\author[1]{Hao-Yun Hsu}
\author[1]{Yi-Ching Cheng}
\author[,1]{Guan-Hua Huang\thanks{Corresponding author: ghuang@nycu.edu.tw}}
\affil[1]{Institute of Statistics, National Yang Ming Chiao Tung University, Hsinchu, Taiwan}
\date{}

\begin{document}

\maketitle

\bigskip

\begin{abstract}
\noindent This study explored the architecture of semantic segmentation and evaluated models that excel in polyp segmentation. We present an integrated framework that harnesses the advantages of different models to attain an optimal outcome. Specifically, in this framework, we fuse the learned features from convolutional and transformer models for prediction, thus engendering an ensemble technique to enhance model performance. Our experiments on polyp segmentation revealed that the proposed architecture surpassed other top models, exhibiting improved learning capacity and resilience. The code is available at \url{https://github.com/HuangDLab/EnFormer}.
\end{abstract}


\section{Introduction}

Polyp segmentation is a critical task in medical image analysis, particularly in the context of gastrointestinal health. Polyps are abnormal tissue growths that can develop on the lining of the colon or rectum. Although many polyps are benign, some can evolve into colorectal cancer. Therefore, early detection and removal are crucial for preventing cancer progression.

In recent years, convolutional neural networks (CNNs) \citep{resnet, densenet, mobilenet, efficientnet, efficientnetv2} and transformer architectures \citep{vit, swin, pvt, pvtv2, coat} have demonstrated promise in the field of medical image segmentation. These models can learn complex features from training data, enabling them to generalize well to new, unseen images. Moreover, CNNs have been widely integrated with transformer architectures for semantic segmentation \citep{hybrid0, hybrid1, hybrid2}. CNNs and transformers have different assumptions that guide the learning process. CNNs incorporate two inductive biases, namely locality and translation equivariance. Locality equivariance involves the assumption that pixels in close spatial proximity are more likely to be related than those that are further apart, and translation equivariance involves the assumption that a shift in the input image causes a corresponding shift in the feature maps generated by the convolutional layers. These attributes enable convolutional layers to focus on capturing a small portion of the input image rather than the whole image, thus causing CNNs to continue to lead in performance on vision tasks. By contrast, the transformer architecture does not possess these inherent inductive biases. Instead, a transformer must acquire biases from large-scale datasets by leveraging an attention mechanism \citep{attention} to capture long-range dependencies. Thus, transformers tend to excel with large-scale datasets, whereas CNNs typically yield superior results with smaller datasets.

In summary, CNNs excel at capturing local structures, whereas transformers are adept at modeling global information. To create a model that comprehensively understands both local and global patterns, hybrid architectures combining CNNs and transformers have been suggested to leverage the strengths of both architectures. In particular, some studies \citep{fcbformer, fcb-swinv2, transfuse} have combined a CNN framework with a transformer framework; the CNN framework is used to capture the edge and corner features of images, and the transformer framework is used to capture long-distance relationships. We consider this approach as an ensemble approach that integrates the strengths of different architectures to determine a final solution. These architectures have achieved considerable success in computer vision applications, including in polyp segmentation. Conventional polyp segmentation methods are based on Unet-like \citep{unet} CNN architectures, which act as the foundational model. However, as numerous transformer architectures are introduced, research has shifted toward exploring the limitations of using transformers as the core component. This shift has led to substantial improvements relative to CNN architectures.

To the best of our knowledge, studies have yet to experiment with different decoding strategies for polyp segmentation. Accordingly, this study comprehensively examined various decoding strategies that have achieved considerable success in computer vision but have not been investigated for polyp segmentation. Additionally, we explored ensemble techniques utilized in polyp segmentation and reformulated these techniques to establish a more general architecture of integrating different encoders' multiscale feature maps to enhance model performance. We conducted experiments on five well-known datasets: Kvasir \citep{Kvasir-SEG}, CVC-ClinicDB \citep{CVC-ClinicDB}, CVC-ColonDB \citep{CVC-ColonDB}, CVC-300 \citep{CVC-300}, and ETIS-LaribPolypDB \citep{ETIS}. The first two datasets were used to assess the model's learning ability, and the remaining three were used to evaluate the model's generalization ability. The final experimental results indicated that our model demonstrated greater learning and robustness ability compared with other models.

\section{Related work}

\subsection{Semantic segmentation}

Semantic segmentation models typically comprise two main components: an encoder and a decoder. The encoder's role is to extract semantic details at various scales from the input image, and the decoder uses this information to produce the final output. In encoder design, the ResNet architecture \citep{resnet, resnext, resnest} is the most commonly used convolution module for feature extraction. It passes the input through one or more layers for residual mapping to help mitigate the issue of gradient vanishing. In the context of transformer backbones, a study proposed the Pyramid Vision Transformer (PVT) \citep{pvt} that involves a progressively shrinking pyramid and spatial-reduction attention layer to create high-resolution, multiscale feature maps, enabling Vision Transformer (ViT) \citep{vit} models to perform effectively in dense prediction. Furthermore, an enhanced version of the original PVT, namely PVTv2 \citep{pvtv2}, was proposed by adding three modules: a linear complexity attention layer, overlapping patch embedding, and a convolutional feed-forward network. CoaT \citep{coat}, another transformer backbone, integrates coscale and conv-attentional mechanisms, demonstrating superior performance compared with the PVT.

In the design of semantic segmentation models, the decoder component seeks to restore the original image dimensions from multiscale feature maps. Typically, the decoder employs upsampling techniques and a progression of matrix operations. Traditional Unet \citep{unet} architectures use skip connections to combine feature maps from the encoder into the decoder, leading to successful results. A study developed a UNet++ architecture \citep{unet++} by redesigning  the skip connections and introducing deep supervision to average the feature maps created by each segmentation branch. Moreover, studies have reported that alternative decoder designs, such as those used by DeepLab \citep{deeplabv3, deeplabv3+}, achieved impressive performance on benchmark datasets such as MS COCO \citep{coco} and VOC \citep{voc}. DeepLab uses the Atrous Spatial Pyramid Pooling (ASPP) block to effectively capture multilevel information for both small and large objects by dilating the receptive fields, leading to state-of-the-art performance. On the basis of Unet, an MA-Net architecture \citep{manet} was designed using a self-attention mechanism to adaptively combine local features with their global contexts. A\ study also introduced a LinkNet architecture \citep{linknet} that adheres to the standard encoder-decoder architecture commonly used in segmentation tasks; in this architecture, a new method is introduced for connecting each encoder with its corresponding decoder. The Feature Pyramid Network (FPN) architecture \citep{fpn} leverages attention mechanisms in conjunction with spatial pyramids to achieve precise dense feature extraction for pixel-wise classification, avoiding the complexity of dilated convolutions and manually designed decoder networks. The Pyramid Scene Parsing Network (PSPNet) architecture \citep{pspnet} includes a novel pyramid pooling module to aggregate context from different regions. The Pyramid Attention Network (PAN) architecture \citep{pan} improves semantic segmentation by merging attention mechanisms with spatial pyramids to accurately extract dense features for pixel labeling. This network includes a Feature Pyramid Attention module for better feature representation and a Global Attention Upsample module to assist in low-level feature localization. Other architectures involve various strategies for setting up encoders and decoders as well as different approaches for integrating features from the encoder into the decoder.

\subsection{Polyp segmentation}

In polyp segmentation, a Unet-like architecture is often selected as a baseline model. To enhance the performance of segmentation models, researchers have focused on devising more precise and efficient methods for decoding the feature maps created by the input image. Some of these methods are described as follows:\ PraNet \citep{pranet} leverages a parallel partial decoder to generate a high-level semantic global map and employs multiple reverse attention modules for accurate polyp segmentation. Polyp-PVT \citep{polyppvt}, which uses a transformer encoder and incorporates difference modules to collect the semantic and location information from the low-level and high-level features, significantly enhances polyp segmentation by effectively merging cross-level features and reducing noise, thus surpassing CNN-based methods in various difficult scenarios.

Scholars have extensively integrated CNN and transformer architectures for polyp segmentation because this integration can leverage the strengths of these two distinct encoders for feature extraction. For example, a study proposes an integrated model, namely TransFuse \citep{transfuse}, that derives data features by processing the data through both a convolution-based encoder and a transformer-based encoder, subsequently determining the model output based on the features extracted by these dual encoders. In particular, TransFuse uses ResNet as the convolutional encoder and DeiT \citep{deit} as the transformer encoder for feature extraction. The features extracted by both the convolutional and transformer encoders are then combined into a comprehensive feature vector. This approach can be viewed as a type of ensemble model that produces the final model output by combining the outputs of the convolutional and transformer models. Another study proposed an ensemble model, namely  FCBFormer \citep{fcbformer}, that has two branches, namely convolution and transformer branches. Each branch includes an encoder and a decoder architecture, and the final prediction is generated by incorporating the decoded features of these two branches and passing them through a small neural network.

\subsection{Architecture in FCBFormer}

We consider FCBFormer as a stacking approach \citep{stacking} in an ensemble technique that combines the outputs from both the convolution branch and the transformer branch. Specifically, the convolutional branch in FCBFormer employs a combination of Residual Blocks (RBs) with both downsampling and upsampling components. The link between the encoder and decoder functions similarly to that in a Unet structure. Let the encoder block and decoder block of the convolutional branch be CB$_E$ and CB$_D,$ respectively. For an input $x$, an RB can be defined as follows:
\begin{align*}
    \mathrm{GSC}(x) &= \mathrm{Conv}(\mathrm{SiLU}(\mathrm{GN}(x))) \\
    \mathrm{RB}(x) &= x + \mathrm{GSC}(\mathrm{GSC}(x))
\end{align*}
\noindent where $\mathrm{GN}$ denotes group normalization, $\mathrm{SiLU}$ denotes a nonlinear activation function, and $\mathrm{Conv}$ denotes  a standard convolutional layer. In the transformer branch, FCBFormer employs PVTv2-B3 \citep{pvtv2} as the encoder to derive feature maps. These extracted features are subsequently processed by a local emphasis (LE) module to improve local feature representations; subsequently, multiscale features are integrated through a stepwise feature aggregation (SFA) module. The LE and SFA modules are defined as follows:
\begin{align*}
    \mathrm{RB}^2(x) &= \mathrm{RB}(\mathrm{RB}(x)) \\
    \mathrm{LE}(x) &= \mathrm{Up}_{4}(\mathrm{RB}^2(x)) \\
    \mathrm{D}(x, y) &= \mathrm{RB}^2(\mathrm{concat}(x, y)) \\
    \mathrm{SFA}(e^1, e^2, e^3, e^4) &= \mathrm{D}(\mathrm{D}(\mathrm{D}(e^4, e^3), e^2), e^1)
\end{align*}
\noindent where $\mathrm{RB}^2$ represents double RBs and $\mathrm{Up}_s$ represents an upsampling block that takes input feature maps and upsamples them to $(\frac{H}{s}, \frac{W}{s})$. Moreover, $\mathrm{D}$ represents a decoder block that concatenates the output of the preceding decoder block with the skip connection from the transformer encoder. Let $e^j$ be the LE-improved extracted feature map from the $j$th stage of the transformer encoder. The entire SFA and LE processes are collectively termed as the Improved Progressive Locality Decoder (PLD+) for enhancing the transformer's capability in extracting local features. Finally, the prediction head (PH) generates the final predicted probabilities:
\begin{align*}
    \mathrm{PH}(x) &= \sigma(\mathrm{Conv}(\mathrm{RB}^2(\mathrm{Up}_1(x))))
\end{align*}
\noindent The input to the PH is formed by combining the outputs of the convolutional and transformer branches. This combined output is then resized to match the original image dimensions, passed through $\mathrm{RB}^2$ to refine the decoding feature, and finally processed by a convolutional layer with a kernel size of 1, followed by a sigmoid nonlinear transformation ($\sigma$) for the final prediction.

\section{Methods}

Semantic segmentation involves three main sequential stages: encoding, decoding, and prediction. The encoding stage, represented by $\mathcal{E}$, processes an input image to extract multiscale information. Specifically, the $j$th block of the encoder $\mathcal{E}$ is denoted as $\mathcal{E}^j$. The decoding stage, denoted as $\mathcal{D}$, then uses the feature maps generated at the encoding stage to derive more precise features. Finally, the output of $\mathcal{D}$ is fed into a segmentation head $\mathcal{S}$ to segment objects in the image. For notational simplicity, these stages are combined into a composite function $\mathcal{H} = \mathcal{S} \circ \mathcal{D} \circ \mathcal{E}$.

In a segmentation task, we consider an image $X$ with dimensions $X\in \mathbb{R} ^ {I \times H\times W}$, where $I$ denotes the number of input channels. The associated label is $Y\in \mathbb{R} ^ {H\times W}$, with $y_{h, w}\in\{0,1,\ldots,C\}$, where $C$ represents the number of segmentation classes. The feature maps produced after processing through the encoder block $\mathcal{E}^j$ are represented as $e^j\in \mathbb{R}^{I^j\times H^j \times W^j}$, where $I^j$, $H^j$, and $W^j$ denote the number of output channels, height of feature maps from $\mathcal{E}^j$, and width of feature maps from $\mathcal{E}^j$, respectively. The decoder $\mathcal{D}$ takes inputs from $\mathcal{E}$ and generates the decoding features $d=\mathcal{D}(\mathcal{E}(X))$. The predicted probabilities are denoted as $P=\mathcal{H}(X)=\mathcal{S}(d)$, where $P\in \mathbb{R} ^ {C \times H\times W}$, and $p_{c,h,w}$ signifies the probability that the input belongs to class $c$ at pixel $(h,w)$.


In an encoder-decoder architecture, the encoder and decoding strategy are essential for performance. For example, ResNet can be used to extract image features, and a Unet decoding strategy can be used to process these multiscale features. Here, $\mathcal{E}=\mathrm{ResNet}$, $\mathcal{E}^j, j=1,\cdots,4$ denotes the feature maps obtained from various stages of ResNet, $\mathcal{D}=\mathrm{Unet}$, and $\mathcal{S}$ denotes a series of convolutional layers followed by an upsampling layer. This notation can be applied to any encoder-decoder architecture.

To extend these notations into ensemble-based approaches, assume that we have $K$ encoders $\mathcal{E}_{1}, \cdots, \mathcal{E}_{K}$. Let $\mathcal{E}^j_i$ represent the $j$th block of the $i$th encoder and $D_i$ represent the $i$th decoder w.r.t. $\mathcal{E}_i$. Let $e_i^j$ be the encoding features and $d_i$ be the decoding features. We also define a module named fusion decoder (FD), denoted as $\mathcal{F}$, to fuse the features generated by different encoders; that is, the input of $\mathcal{F}$ is $e_i^j$'s.

Various ensemble methods have been proposed. One of these methods is a stacking approach, which entails concatenating the outputs of multiple decoders $\mathcal{D}_1,\cdots,\mathcal{D}_K$ and then constructing a segmentation head $\mathcal{S}$ to integrate the information from these decoders for the final prediction. Another method involves fusing feature maps produced in the early stages; the features $e_i^j$ are fused by fusion decoder $\mathcal{F}$, and the output of $\mathcal{F}$ serves as the input for the segmentation head $\mathcal{S}$. For example, in FCBFormer, $K=2$, $\mathcal{E}_1=\mathrm{CB}_E$ serves as the encoder for the convolution branch, and $\mathcal{D}_1=\mathrm{CB}_D$ serves as the decoder. Additionally, $\mathcal{E}_2=$PVTv2-B3 represents the encoder for the transformer branch, and $\mathcal{D}_2=\mathrm{PLD+}$ is its decoder. The final predicted probabilities are computed as $P=\mathcal{H}(X)=\mathcal{S}(d_1,d_2)$, with $\mathcal{S}=\mathrm{PH}$. TransFuse integrates ResNet and DeiT using a BiFusion  architecture \citep{transfuse}. Because ResNet and DeiT lack decoders, in this scenario, $\mathcal{E}_1=\mathrm{ResNet}$, $\mathcal{E}_2=\mathrm{DeiT}$, $\mathcal{F}=\mathrm{BiFusion,}$ and the prediction is given by $P=\mathcal{H}(X)=\mathcal{S}(\mathcal{F}([e_i^j]))$.

\begin{figure}[h]
  \includegraphics[scale=0.09]{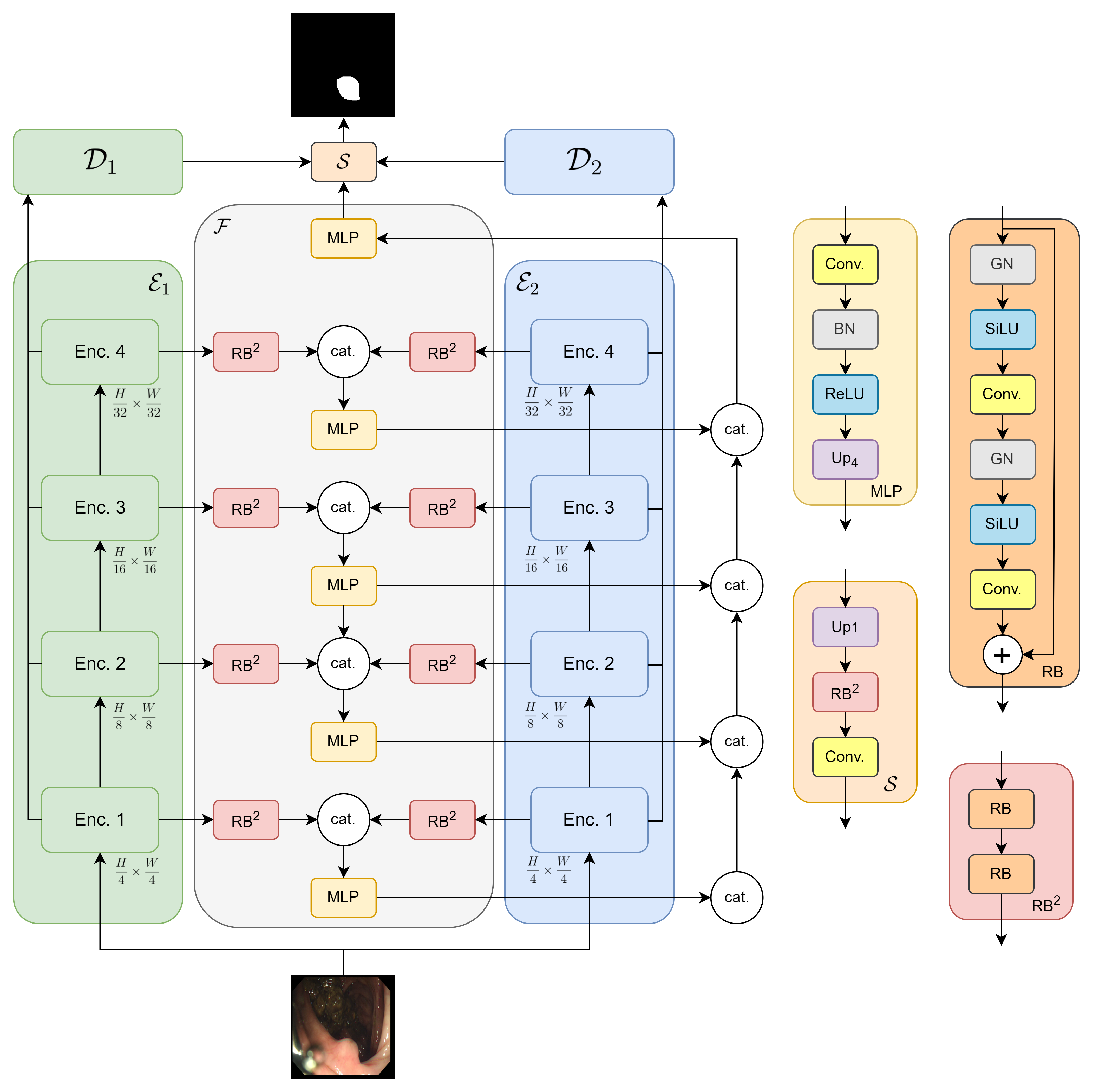}
   \centering
  \caption[EnFormer]{EnFormer. Green and blue areas represent the different encoder configurations for the convolution branch and the transformer branch, respectively. Gray indicates the FD, which combines features generated by both the convolution and transformer branches.}
  \label{fig:enformer}
\end{figure}

\begin{figure}[h]
  \includegraphics[scale=0.09]{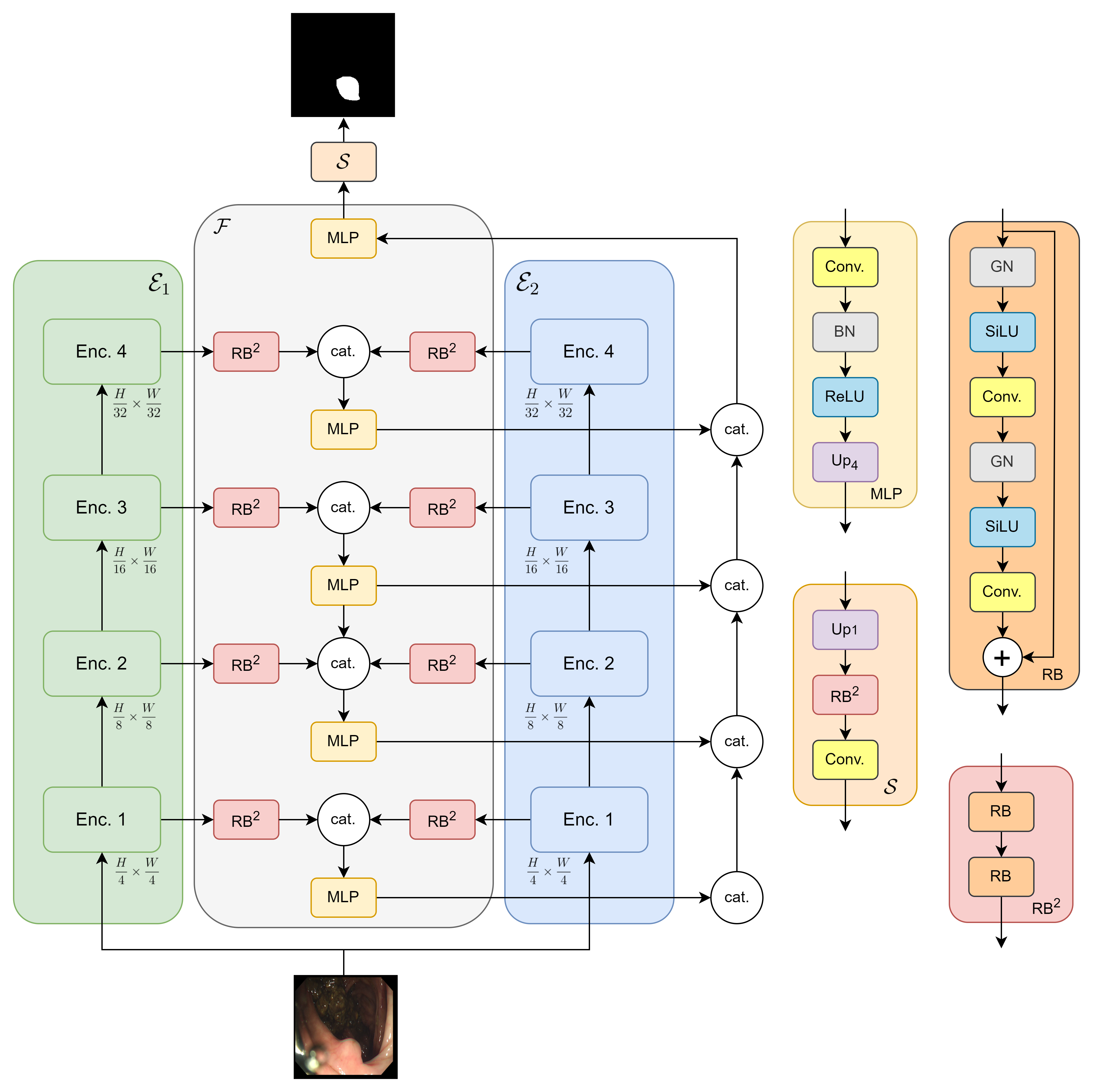}
  \centering
  \caption[EnFormer-Lite]{EnFormer-Lite. The architecture is similar to EnFormer; however, it lacks a decoding strategy for each encoder and is thus a lighter version of EnFormer.}
  \label{fig:enformer-lite}
\end{figure}

\subsection{Proposed method}

Our methodology draws considerable inspiration from FCBFormer. However, we argue that FCBFormer's model architecture is deficient in facilitating information exchange during backpropagation. To mitigate this, we suggest incorporating $\mathcal{F}$ into FCBFormer to enhance information transfer between encoders. Furthermore, owing to constraints related to hardware and computational efficiency considerations, we use only two encoders ($K=2$), and convolution and transformer modules are integrated into our technique to more effectively capture both local and global contexts. Our FD is characterized as follows:
\begin{align*}
    \mathrm{MLP}(x) &= \mathrm{Up}_4(\mathrm{ReLU}(\mathrm{BN}(\mathrm{Conv}(x)))) \\
    \mathcal{F}^j(e_1^j, e_2^j) &= \mathrm{MLP}(\mathrm{concat}(\mathrm{RB}^2(e_1^j), \mathrm{RB}^2(e_2^j))) \\
    \mathcal{F}([e_i^j]) &= \mathrm{MLP}(\mathrm{concat}(\mathcal{F}^j(e_1^j, e_2^j) \ j=1,2,3,4))
\end{align*}
\noindent where BN \citep{batchnorm} denotes batch normalization and ReLU \citep{relu} denotes the nonlinear activation function. The architecture of $\mathcal{F}$ is relatively simple. First, features are extracted using the convolutional encoder $\mathcal{E}_1$ and the transformer encoder $\mathcal{E}_2$. These features are then concatenated and processed through a basic convolution module, $\mathrm{MLP}$, to merge them. The fused features are then fed into $\mathcal{F}$.

Regarding the FD configuration, only the input features generated by the encoders are used. We propose two methods for obtaining the final probability predictions. The first method preserves the decoding strategy for each branch, resulting in the final prediction $P = \mathcal{H}(X) = \mathcal{S}(d_1, d_2, \mathcal{F}([e^j_i]))$, where $\mathcal{S} = \mathrm{PH}$. This method is denoted as Ensemble Transformer and Convolution (EnFormer) and is displayed in Figure \ref{fig:enformer}. The second method involves excluding the decoder configuration in both the convolution and transformer branches, implying that decoding would not occur for either branch. Only the FD $\mathcal{F}$ is used to generate the final probability predictions. In this case, $\mathcal{D}_1 = \mathcal{D}_2 = \mathrm{Null}$, and $P = \mathcal{H}(X) = \mathcal{S}(\mathcal{F}([e^j_i]))$, where $\mathcal{S} = \mathrm{PH}$. This method is denoted as EnFormer-Lite and is illustrated in Figure \ref{fig:enformer-lite}.

In this paper, the FD $\mathcal{F}$ is attached to FCBFormer in the EnFormer architecture; hence, this architecture is directly comparable to FCBFormer. For EnFormer-Lite, we propose four model sizes, namely EnFormer-Lite Mini, EnFormer-Lite Small, and EnFormer-Lite Medium, which use the same convolution branch as FCBFormer but different transformer branches, namely CoaT-Lite Mini, CoaT-Lite Small, and CoaT-Lite Medium, respectively. Finally, EnFormer-Lite Large uses ResNet50 as its convolution branch and CoaT-Lite Medium as the transformer backbones. The models are compared in Table \ref{tab:models}.

\begin{table}[h]
\tabcolsep=5pt
\centering
\caption{Model comparison.}
\label{tab:models}
\begin{tabular}{l|llllll}
\toprule
model & $\mathcal{E}_1$ & $\mathcal{E}_2$ & $\mathcal{D}_1$ & $\mathcal{D}_2$ & $\mathcal{F}$ & $\mathcal{S}$ \\
\midrule
Unet++               & ResNet50 & & Unet++     & & & Conv\\
Unet                 & ResNet50 & & Unet       & & & Conv\\
MAnet                & ResNet50 & & MAnet      & & & Conv\\
Linknet              & ResNet50 & & Linknet    & & & Conv\\
FPN                  & ResNet50 & & FPN        & & & Conv\\
PSPNet               & ResNet50 & & PSPNet     & & & Conv\\
PAN                  & ResNet50 & & PAN        & & & Conv\\
DeepLabV3            & ResNet50 & & DeepLabV3  & & & Conv\\
DeepLabV3+           & ResNet50 & & DeepLabV3+ & & & Conv\\
\bottomrule
FCBFormer            & CB$_E$ & PVTv2-B3 & CB$_D$ & PLD+ & & PH\\
EnFormer             & CB$_E$ & PVTv2-B3 & CB$_D$ & PLD+ & FD & PH\\
EnFormer-Lite Mini   & CB$_E$ & CoaT-Lite Mini & & & FD & PH\\
EnFormer-Lite Small  & CB$_E$ & CoaT-Lite Small & & & FD & PH\\
EnFormer-Lite Medium & CB$_E$ & CoaT-Lite Medium & & & FD & PH\\
EnFormer-Lite Large  & ResNet50 & CoaT-Lite Medium & & & FD & PH\\
\bottomrule
\end{tabular}
\end{table}

\section{Experiments}

\subsection{Dataset}

We conducted experiments to assess the performance of the various models on five datasets: Kvasir \cite{Kvasir-SEG}, CVC-ClinicDB \citep{CVC-ClinicDB}, CVC-300 \citep{CVC-300}, CVC-ColonDB \citep{CVC-ColonDB}, and ETIS-LaribPolypDB \citep{ETIS}. We adopted the data split from \cite{pranet} as the training and test sets for these datasets. The number of images in each dataset is listed in Table \ref{tab:polyp}. However, because \cite{pranet} did not provide a validation set for hyperparameter tuning, we further divided the training  set in a 9:1 ratio to avoid overfitting on the test set. We employed the training sets of the first two datasets for model training. During the testing phase, the test sets of Kvasir and CVC-ClinicDB were used to evaluate the model's learning ability, and those of the remaining three datasets were used to evaluate the model's generalization capability.

The datasets are intended for polyp segmentation tasks, with an emphasis on outlining the polyp areas within the images. Thus, they have only one class ($C=1$). To accelerate the model's training, we resized the original images to 352$\times$352 pixels. To increase dataset diversity, we employed various data augmentation techniques on the basis of previous research  \citep{fcbformer, fcb-swinv2}. For geometric augmentations, we applied multiple transformations to both images and masks, such as horizontal and vertical flips, affine transformations, and grid distortions (dividing the image into a grid of cells and randomly shifting the grid intersections to create localized distortions). Additionally, we used pixel-level color variations such as color jitter and unsharp (using a Gaussian kernel for image filtering) for the images. We also normalized the images by using the standard ImageNet \citep{imagenet} mean and standard deviation. Each augmentation was applied independently with a probability of 50\%.

\begin{table}[h]
  \centering
  \caption{Polyp dataset information.}
  \label{tab:polyp}
  \begin{tabular}{ccccccc}
    \toprule
    \multicolumn{2}{c}{\textbf{Train}} & \multicolumn{5}{c}{\textbf{Test}} \\
    \cmidrule(lr){1-2} \cmidrule(lr){3-7}
    Kvasir & CVC-ClinicDB & Kvasir & CVC-ClinicDB & CVC-300 & CVC-ColonDB & ETIS-LaribPolypDB \\
    \midrule
    900 & 550 & 100 & 62 & 60 & 380 & 196 \\
    \bottomrule
  \end{tabular}
\end{table}

\subsection{Training configuration}

We used the AdamW \citep{adamw} optimizer with a learning rate of 0.0001 and no weight decay. Additionally, we adopted the OneCycleLR \citep{onecyclelr} schedule to regulate the learning rate. To ensure a fair evaluation of the different models, we trained each model for a total of 200 epochs with a batch size of 16. Moreover, the models' encoder parameters were initialized with pretrained weights from ImageNet. The entire training dataset comprised 1305 examples, and the best model was saved based on the highest Dice coefficient on a validation set of 145 images. The loss function for updating the parameters was the average of the Dice loss and binary cross entropy loss. All experiments were conducted using four Tesla V100-SXM2-16GB GPUs. We divided the experiment into three segments. First, we examined the performance of the encoder-decoder model by using nine different decoding strategies with ResNet50 as the encoder. These models were all constructed using the Python package segmentation\_models\_pytorch \citep{smp}. Second, we explored the ensemble-based approach of our proposed EnFormer and EnFormer-Lite models. Finally, we compared our models with FCBFormer, PraNet, and Polyp-PVT. Each of our proposed models was trained using our specified parameters, whereas PraNet and Polyp-PVT were trained with their released code. The best model was chosen on the basis of our validation set partition.

\subsection{Evaluation metrics}

Studies have employed different methods to assess model performance on a test set. For example, the study that proposed FCBFormer assessed images at a resolution of 352 $\times$ 352 by using a 0.5 threshold (i.e., probabilities $>$ 0.5 were classified as polyp areas). By contrast, other studies \citep{pranet, polyppvt, uacanet} have assessed images at their original sizes by using varying thresholds and then computed the average metrics across all thresholds. In accordance with the approach used in \citep{pranet}, we resized the predicted probabilities to each test image's original size and computed the metrics over 256 thresholds, which were evenly distributed within [0,1], for each image in the test set. We then averaged these metrics across all images in the test set. Six metrics were employed for model evaluation: Dice coefficient, intersection over union (IoU), mean absolute error (MAE), weighted F-measure ($F_\beta^w$) \citep{weighted-f-measure}, S-measure ($S_\alpha$) \citep{s-measure}, and E-measure ($E_{\xi}$) \citep{e-measure}. The Dice coefficient and IoU are measures of shape similarity between the predicted mask and the ground truth, and the MAE represents the L1 distance between the predicted probabilities and the ground truth. $F_\beta^w$ is a weighted average of recall and precision, and it is an extension of the F-measure (F1 score) to evaluate pixel-wise accuracy. $S_\alpha$ is a measure of structural similarity, and $E_{\xi}$ is a measure of both local and global similarity.

\subsection{Result analysis}

Our experimental results for the five datasets are listed in Table \ref{tab:kvasir}, \ref{tab:cvc-clinicdb}, \ref{tab:cvc-300}, \ref{tab:cvc-colondb}, and \ref{tab:etis}. The first two datasets were used to assess the model's training performance. Their training images weer used in the training phase, and the validation images were employed to choose the best model for evaluation on the test set. However, no images in the last three datasets were included in the training phase; these images were instead used to evaluate the model's generalization ability. From the first two datasets, our EnFormer-Lite Large achieved the highest Dice coefficient on Kvasir, and EnFormer achieved the highest Dice coefficient on CVC-ClinicDB. However, FCBFormer also had excellent performance on both datasets. Additionally, the encoder-decoder model yielded remarkable results and was competitive with the ensemble model. By contrast, for the last three datasets, the encoder-decoder model had substantially lower performance than did the ensemble model, indicating that the ensemble model is more robust. In our experiment, our model had greater robustness than did FCBFormer, suggesting that integrating the FD can enhance robustness without compromising learning ability.

\begin{table}[h]
\tabcolsep=5pt
\centering
\caption{Evaluation metrics on the Kvasir test set.}
\label{tab:kvasir}
\begin{threeparttable}
\begin{tabular}{l|ccccccc}
\toprule
model & mDice \tnote{1} $(\uparrow)$ \tnote{3} & mIoU $(\uparrow)$ & $F_\beta^w$ $(\uparrow)$ & $S_\alpha$ $(\uparrow)$ & $mE_{\xi}$ $(\uparrow)$ & $maxE_{\xi}$ \tnote{2} $(\uparrow)$ & MAE $(\downarrow)$ \tnote{4} \\
\midrule
Unet       & 0.8773 & 0.8222 & 0.8677 & 0.9043 & 0.9300 & 0.9360 & 0.0330 \\
Unet++     & 0.8771 & 0.8220 & 0.8673 & 0.9049 & 0.9319 & 0.9416 & 0.0339 \\
MAnet      & 0.8952 & 0.8414 & 0.8879 & 0.9157 & 0.9468 & 0.9534 & 0.0288 \\
Linknet    & 0.8905 & 0.8320 & 0.8776 & 0.9107 & 0.9427 & 0.9483 & 0.0313 \\
FPN        & 0.8873 & 0.8291 & 0.8778 & 0.9100 & 0.9418 & 0.9498 & 0.0319 \\
PSPNet     & 0.8517 & 0.7738 & 0.8205 & 0.8880 & 0.9159 & 0.9330 & 0.0441 \\
PAN        & 0.8902 & 0.8346 & 0.8828 & 0.9132 & 0.9405 & 0.9494 & 0.0277 \\
DeepLabV3  & 0.8908 & 0.8325 & 0.8823 & 0.9098 & 0.9450 & 0.9496 & 0.0286 \\
DeepLabV3+ & 0.8827 & 0.8223 & 0.8676 & 0.9076 & 0.9376 & 0.9444 & 0.0325 \\
\bottomrule
Polyp-PVT  & 0.8973 & 0.8449 & 0.8887 & 0.9117 & 0.9486 & 0.9524 & 0.0283 \\
PraNet    & 0.8840 & 0.8253 & 0.8692 & 0.9054 & 0.9304 & 0.9346 & 0.0332 \\
FCBFormer & 0.9129 & 0.8611 & 0.9045 & 0.9257 & 0.9523 & 0.9565 & 0.0245 \\
\bottomrule
EnFormer-Lite Mini   & 0.8974 & 0.8386 & 0.8839 & 0.9145 & 0.9453 & 0.9527 & 0.0301 \\
EnFormer-Lite Small  & 0.9113 & 0.8615 & 0.9045 & 0.9280 & 0.9569 & 0.9622 & 0.0248 \\
EnFormer-Lite Medium & 0.9098 & 0.8588 & 0.9066 & 0.9307 & 0.9486 & 0.9559 & 0.0260 \\
EnFormer-Lite Large  & \textbf{0.9224} & \textbf{0.8731} & \textbf{0.9167} & \textbf{0.9359} & \textbf{0.9605} & \textbf{0.9652} & \textbf{0.0228} \\
EnFormer             & 0.9059 & 0.8526 & 0.8994 & 0.9227 & 0.9500 & 0.9556 & 0.0268 \\
\bottomrule
\end{tabular}
\begin{tablenotes}
$ ^1$ ``m" means the average across the test set.
$ ^2$ ``max" represents the maximum values across the test set.
$ ^3$ $\uparrow$ means a higher value is better.
$ ^4$ $\downarrow$ means a lower value is better.
Values in boldface denote the best result for each evaluation metric.
\end{tablenotes}
\end{threeparttable}
\end{table}

\begin{table}[H]
\tabcolsep=5pt
\centering
\caption{Evaluation metrics on the CVC-ClinicDB test set.}
\label{tab:cvc-clinicdb}
\begin{threeparttable}
\begin{tabular}{l|ccccccc}
\toprule
model & mDice \tnote{1} $(\uparrow)$ \tnote{3} & mIoU $(\uparrow)$ & $F_\beta^w$ $(\uparrow)$ & $S_\alpha$ $(\uparrow)$ & $mE_{\xi}$ $(\uparrow)$ & $maxE_{\xi}$ \tnote{2} $(\uparrow)$ & MAE $(\downarrow)$ \tnote{4} \\
\midrule
Unet                 & 0.8881 & 0.8350 & 0.8815 & 0.9295 & 0.9508 & 0.9622 & 0.0108 \\
Unet++               & 0.9057 & 0.8512 & 0.8964 & 0.9352 & 0.9654 & 0.9699 & 0.0108 \\
MAnet                & 0.9052 & 0.8522 & 0.8960 & 0.9359 & 0.9613 & 0.9665 & 0.0103 \\
Linknet              & 0.9079 & 0.8543 & 0.8981 & 0.9356 & 0.9665 & 0.9769 & 0.0098 \\
FPN                  & 0.9003 & 0.8443 & 0.8928 & 0.9383 & 0.9606 & 0.9685 & 0.0122 \\
PSPNet               & 0.8694 & 0.8005 & 0.8540 & 0.9135 & 0.9498 & 0.9651 & 0.0195 \\
PAN                  & 0.9010 & 0.8482 & 0.8965 & 0.9363 & 0.9594 & 0.9763 & 0.0119 \\
DeepLabV3            & 0.9213 & 0.8651 & 0.9171 & 0.9459 & 0.9783 & 0.9856 & 0.0089 \\
DeepLabV3+           & 0.8969 & 0.8432 & 0.8837 & 0.9381 & 0.9600 & 0.9690 & 0.0105 \\
\bottomrule
PraNet               & 0.9174 & 0.8646 & 0.9120 & 0.9465 & 0.9703 & 0.9751 & 0.0107 \\
Polyp-PVT            & 0.9045 & 0.8503 & 0.9020 & 0.9307 & 0.9594 & 0.9657 & 0.0137 \\
FCBFormer            & 0.9369 & 0.8899 & 0.9334 & \textbf{0.9571} & 0.9829 & 0.9878 & \textbf{0.0064} \\
\bottomrule
EnFormer-Lite Mini   & 0.9162 & 0.8588 & 0.9086 & 0.9405 & 0.9753 & 0.9810 & 0.0099 \\
EnFormer-Lite Small  & 0.9171 & 0.8634 & 0.9171 & 0.9441 & 0.9766 & 0.9861 & 0.0110 \\
EnFormer-Lite Medium & 0.9304 & 0.8792 & 0.9243 & 0.9538 & 0.9816 & 0.9885 & 0.0083 \\
EnFormer-Lite Large  & 0.9348 & 0.8875 & 0.9319 & 0.9532 & 0.9802 & 0.9849 & 0.0072 \\
EnFormer             & \textbf{0.9390} & \textbf{0.8914} & \textbf{0.9381} & 0.9530 & \textbf{0.9848} & \textbf{0.9891} & 0.0073 \\
\bottomrule
\end{tabular}
\begin{tablenotes}
$ ^1$ ``m" means the average across the test set.
$ ^2$ ``max" represents the maximum values across the test set.
$ ^3$ $\uparrow$ means a higher value is better.
$ ^4$ $\downarrow$ means a lower value is better.
Values in boldface denote the best result for each evaluation metric.
\end{tablenotes}
\end{threeparttable}
\end{table}

\begin{table}[H]
\tabcolsep=5pt
\centering
\caption{Evaluation metrics on CVC-300.}
\label{tab:cvc-300}
\begin{threeparttable}
\begin{tabular}{l|ccccccc}
\toprule
model & mDice \tnote{1} $(\uparrow)$ \tnote{3} & mIoU $(\uparrow)$ & $F_\beta^w$ $(\uparrow)$ & $S_\alpha$ $(\uparrow)$ & $mE_{\xi}$ $(\uparrow)$ & $maxE_{\xi}$ \tnote{2} $(\uparrow)$ & MAE $(\downarrow)$ \tnote{4} \\
\midrule
Unet & 0.8771 & 0.8153 & 0.8634 & 0.9279 & 0.9554 & 0.9627 & 0.0072 \\
Unet++ & 0.8580 & 0.7848 & 0.8325 & 0.9168 & 0.9386 & 0.9465 & 0.0096 \\
MAnet & 0.8528 & 0.7924 & 0.8419 & 0.9224 & 0.9253 & 0.9348 & 0.0120 \\
Linknet & 0.8599 & 0.7837 & 0.8265 & 0.9142 & 0.9425 & 0.9549 & 0.0096 \\
FPN & 0.8884 & 0.8188 & 0.8619 & 0.9341 & 0.9598 & 0.9851 & 0.0090 \\
PSPNet & 0.8027 & 0.7054 & 0.7459 & 0.8830 & 0.9052 & 0.9333 & 0.0160 \\
PAN & 0.8978 & 0.8278 & 0.8781 & 0.9384 & 0.9712 & 0.9812 & 0.0061 \\
DeepLabV3 & 0.8639 & 0.7994 & 0.8489 & 0.9261 & 0.9378 & 0.9442 & 0.0087 \\
DeepLabV3+ & 0.8528 & 0.7835 & 0.8232 & 0.9215 & 0.9284 & 0.9605 & 0.0120 \\
\bottomrule
PraNet & 0.8924 & 0.8210 & 0.8705 & 0.9349 & 0.9680 & 0.9817 & 0.0067 \\
Polyp-PVT & 0.8822 & 0.8089 & 0.8599 & 0.9262 & 0.9584 & 0.9696 & 0.0100 \\
FCBFormer & 0.8786 & 0.8064 & 0.8518 & 0.9269 & 0.9527 & 0.9703 & 0.0093 \\
\bottomrule
EnFormer-Lite Mini & 0.8796 & 0.8048 & 0.8578 & 0.9274 & 0.9594 & 0.9657 & 0.0069 \\
EnFormer-Lite Small & \textbf{0.9091} & \textbf{0.8458} & \textbf{0.8924} & \textbf{0.9467} & \textbf{0.9765} & \textbf{0.9874} & \textbf{0.0058} \\
EnFormer-Lite Medium & 0.8865 & 0.8195 & 0.8627 & 0.9334 & 0.9543 & 0.9779 & 0.0087 \\
EnFormer-Lite Large & 0.8897 & 0.8245 & 0.8701 & 0.9353 & 0.9603 & 0.9715 & 0.0065 \\
EnFormer & 0.8925 & 0.8261 & 0.8727 & 0.9350 & 0.9624 & 0.9798 & 0.0075 \\
\bottomrule
\end{tabular}
\begin{tablenotes}
$ ^1$ ``m" means the average across the test set.
$ ^2$ ``max" represents the maximum values across the test set.
$ ^3$ $\uparrow$ means a higher value is better.
$ ^4$ $\downarrow$ means a lower value is better.
Values in boldface denote the best result for each evaluation metric.
\end{tablenotes}
\end{threeparttable}
\end{table}

\begin{table}[H]
\tabcolsep=5pt
\centering
\caption{Evaluation metrics on CVC-ColonDB.}
\label{tab:cvc-colondb}
\begin{threeparttable}
\begin{tabular}{l|ccccccc}
\toprule
model & mDice \tnote{1} $(\uparrow)$ \tnote{3} & mIoU $(\uparrow)$ & $F_\beta^w$ $(\uparrow)$ & $S_\alpha$ $(\uparrow)$ & $mE_{\xi}$ $(\uparrow)$ & $maxE_{\xi}$ \tnote{2} $(\uparrow)$ & MAE $(\downarrow)$ \tnote{4} \\
\midrule
Unet & 0.6871 & 0.6201 & 0.6802 & 0.8080 & 0.8190 & 0.8608 & 0.0466 \\
Unet++ & 0.6909 & 0.6187 & 0.6825 & 0.8075 & 0.8217 & 0.8609 & 0.0447 \\
MAnet & 0.7368 & 0.6690 & 0.7245 & 0.8321 & 0.8540 & 0.8628 & 0.0448 \\
Linknet & 0.7255 & 0.6462 & 0.7078 & 0.8243 & 0.8504 & 0.8667 & 0.0442 \\
FPN & 0.7370 & 0.6614 & 0.7216 & 0.8337 & 0.8613 & 0.8719 & 0.0436 \\
PSPNet & 0.6370 & 0.5439 & 0.6017 & 0.7711 & 0.8011 & 0.8175 & 0.0528 \\
PAN & 0.7045 & 0.6318 & 0.6982 & 0.8124 & 0.8367 & 0.8808 & 0.0448 \\
DeepLabV3 & 0.7184 & 0.6430 & 0.7071 & 0.8211 & 0.8471 & 0.8797 & 0.0427 \\
DeepLabV3+ & 0.7089 & 0.6309 & 0.6916 & 0.8178 & 0.8367 & 0.8516 & 0.0465 \\
\bottomrule
PraNet & 0.7185 & 0.6458 & 0.7045 & 0.8245 & 0.8564 & 0.8775 & 0.0391 \\
Polyp-PVT & 0.7999 & 0.7192 & 0.7792 & 0.8607 & \textbf{0.9052} & \textbf{0.9099} & \textbf{0.0310} \\
FCBFormer & 0.7834 & 0.7060 & 0.7649 & 0.8556 & 0.8840 & 0.8896 & 0.0360 \\
\bottomrule
EnFormer-Lite Mini & 0.7614 & 0.6752 & 0.7429 & 0.8421 & 0.8835 & 0.8910 & 0.0393 \\
EnFormer-Lite Small & 0.7916 & 0.7159 & 0.7782 & 0.8643 & 0.8985 & 0.9080 & 0.0320 \\
EnFormer-Lite Medium & \textbf{0.8055} & \textbf{0.7246} & 0.7868 & \textbf{0.8726} & 0.9042 & 0.9087 & 0.0314 \\
EnFormer-Lite Large & 0.8019 & 0.7200 & 0.7866 & 0.8669 & 0.9034 & 0.9082 & 0.0317 \\
EnFormer & 0.8034 & 0.7241 & \textbf{0.7881} & 0.8702 & 0.9032 & 0.9087 & 0.0311 \\
\bottomrule
\end{tabular}
\begin{tablenotes}
$ ^1$ ``m" means the average across the test set.
$ ^2$ ``max" represents the maximum values across the test set.
$ ^3$ $\uparrow$ means a higher value is better.
$ ^4$ $\downarrow$ means a lower value is better.
Values in boldface denote the best result for each evaluation metric.
\end{tablenotes}
\end{threeparttable}
\end{table}

\begin{table}[H]
\tabcolsep=5pt
\centering
\caption{Evaluation metrics on ETIS.}
\label{tab:etis}
\begin{threeparttable}
\begin{tabular}{l|ccccccc}
\toprule
model & mDice \tnote{1} $(\uparrow)$ \tnote{3} & mIoU $(\uparrow)$ & $F_\beta^w$ $(\uparrow)$ & $S_\alpha$ $(\uparrow)$ & $mE_{\xi}$ $(\uparrow)$ & $maxE_{\xi}$ \tnote{2} $(\uparrow)$ & MAE $(\downarrow)$ \tnote{4} \\
\midrule
Unet & 0.7536 & 0.6818 & 0.7208 & 0.8652 & 0.8736 & 0.8946 & 0.0116 \\
Unet++ & 0.6460 & 0.5857 & 0.6263 & 0.8067 & 0.8002 & 0.8661 & 0.0169 \\
MAnet & 0.7167 & 0.6487 & 0.6831 & 0.8410 & 0.8616 & 0.8746 & 0.0165 \\
Linknet & 0.6777 & 0.6098 & 0.6437 & 0.8244 & 0.8181 & 0.8585 & 0.0170 \\
FPN & 0.7430 & 0.6635 & 0.6970 & 0.8607 & 0.8804 & 0.8979 & 0.0147 \\
PSPNet & 0.6137 & 0.5145 & 0.5504 & 0.7738 & 0.7797 & 0.8032 & 0.0277 \\
PAN & 0.6866 & 0.6131 & 0.6524 & 0.8258 & 0.8477 & 0.8641 & 0.0156 \\
DeepLabV3 & 0.6866 & 0.6139 & 0.6494 & 0.8257 & 0.8359 & 0.8577 & 0.0186 \\
DeepLabV3+ & 0.6973 & 0.6151 & 0.6541 & 0.8315 & 0.8618 & 0.8785 & 0.0181 \\
\bottomrule
PraNet & 0.6534 & 0.5863 & 0.6284 & 0.8070 & 0.8240 & 0.8640 & 0.0191 \\
Polyp-PVT & 0.7630 & 0.6785 & 0.7141 & 0.8617 & 0.8796 & 0.9028 & 0.0211 \\
FCBFormer & 0.7955 & 0.7154 & 0.7505 & 0.8851 & 0.8958 & 0.9301 & 0.0148 \\
\bottomrule
EnFormer-Lite Mini & 0.7601 & 0.6757 & 0.7118 & 0.8658 & 0.8949 & 0.9116 & 0.0179 \\
EnFormer-Lite Small & 0.8258 & 0.7531 & 0.7963 & 0.9041 & 0.9234 & 0.9317 & 0.0116 \\
EnFormer-Lite Medium & 0.7871 & 0.7126 & 0.7445 & 0.8814 & 0.8935 & 0.9274 & 0.0194 \\
EnFormer-Lite Large & 0.7857 & 0.7031 & 0.7332 & 0.8754 & 0.8913 & 0.9264 & 0.0182 \\
EnFormer & \textbf{0.8406} & \textbf{0.7634} & \textbf{0.8030} & \textbf{0.9068} & \textbf{0.9440} & \textbf{0.9545} & \textbf{0.0109} \\
\bottomrule
\end{tabular}
\begin{tablenotes}
$ ^1$ ``m" means the average across the test set.
$ ^2$ ``max" represents the maximum values across the test set.
$ ^3$ $\uparrow$ means a higher value is better.
$ ^4$ $\downarrow$ means a lower value is better.
Values in boldface denote the best result for each evaluation metric.
\end{tablenotes}
\end{threeparttable}
\end{table}

\begin{figure}[h]
  \includegraphics[width=\textwidth]{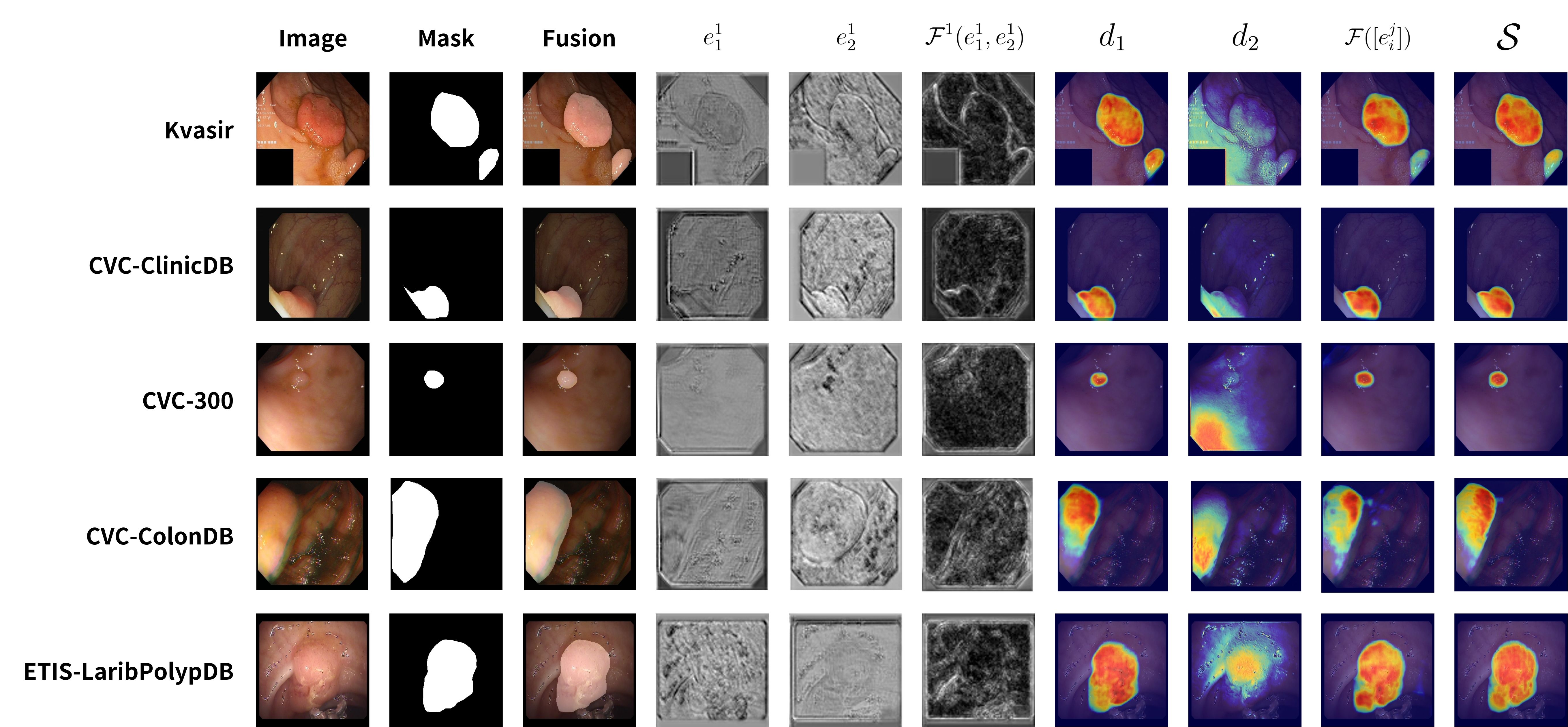}
  \centering
  \caption[Grad-CAM Visualization]{Grad-CAM visualizations for the five datasets. The first three columns (image, mask, and fusion) highlight the regions in the original image identified by the segmentation mask. The columns $e_1^1$, $e_2^1,$ and $\mathcal{F}^1(e_1^1, e_2^1)$ correspond to the average outputs of the first block of the transformer encoder, convolution encoder, and FD, respectively. The columns $d_1$, $d_2,$ and $\mathcal{F}([e^j_i])$ represent the Grad-CAM visualizations of the last layer of the transformer decoder, convolution decoder, and fuse decoder, respectively. Column $\mathcal{S}$ represents the Grad-CAM visualization of the segmentation head.}
  \label{fig:cam}
\end{figure}

\subsection{Visualization}

To explore the effects of the transformer and convolution branches on model outcomes, we created visual representations of the feature maps generated by each branch. We standardized the feature maps of each module (i.e., transformer module and CNN module) to the interval [0,1] and summed the results to emphasize the regions of the image detected by each branch. Additionally, to determine the image regions that the model relies on for its predictions, we applied the basic Grad-CAM \citep{gradcam} visualization method for each dataset. The visual representation for EnFormer is displayed in Figure \ref{fig:cam}.

The visualizations clearly demonstrated that the two encoders captured distinct textures within the images. In the decoding part, the transformer branch and the fuse decoder played key roles in computing the majority of the gradients. The convolution branch occasionally complemented the transformer branch (e.g., in CVC-ColonDB). Furthermore, the segmentation head $\mathcal{S}$ could effectively merge feature maps from various branches. Although convolution branches may occasionally capture incorrect features (e.g., in CVC-300), the fuse decoder could discern which branches accurately detect polyp regions.

\section{Conclusion}

This study explored various decoding strategies and ensemble-based models for polyp segmentation. Our experiments indicated that although encoder-decoder models effectively learned the polyp regions in the training set, they struggled to generalize to unseen images. By contrast, the ensemble-based models significantly outperformed the encoder-decoder models on unseen images, demonstrating higher robustness. Furthermore, we found that selecting an appropriate encoder within the ensemble-based approach is superior to devising an advanced decoding strategy for the features learned by the encoders. For example, our proposed EnFormer-Lite, which includes a simple and lightweight decoder, achieved superior performance in terms of Dice coefficient across the five datasets assessed in the experiments.



\end{document}